# Zenbo Patrol: A Social Assistive Robot Based on Multimodal Deep Learning for Real-time Illegal Parking Recognition and Notification

[1]*Jian-jie Zheng* (鄭建捷), [1]*Chih-kai Yang* (楊智凱), [2]*Po-han Chen* (陳柏翰)*, and* [3,*]*Lyn Chao-ling Chen* (陳昭伶)

[1] Department of Management Information Systems,
[2] Department of Statistics,
[3] Interdisciplinary Artificial Intelligence Center,
National Chengchi University, Taipei City, Taiwan,
E-mail: lynchen@ntu.edu.tw

**ABSTRACT**

In the study, the social robot act as a patrol to recognize and notify illegal parking in real-time. Dual-model pipeline method and large multimodal model were compared, and the GPT-4o multimodal model was adopted in license plate recognition without preprocessing. For moving smoothly on a flat ground, the robot navigated in a simulated parking lot in the experiments. The robot changes angle view of the camera automatically to capture the images around with the format of license plate number. From the captured images of the robot, the numbers on the plate are recognized through the GPT-4o model, and identifies legality of the numbers. When an illegal parking is detected, the robot sends Line messages to the system manager immediately. The contribution of the work is that a novel multimodal deep learning method has validated with high accuracy in license plate recognition, and a social assistive robot is also provided for solving problems in a real scenario, and can be applied in an indoor parking lot.

***Keywords:*** *Line chatbot, Large multimodal model, License plate recognition, Multimodal deep learning, Object detection, Social assistive robot.*

## 1. INTRODUCTION

The issue of campus illegal parking management has considered, and developed the *Zenbo Patrol* (Fig. 1) for license plate recognition and illegal parking notification. For the rapid development of artificial intelligence, the LMMs (Large multimodal models) has been integrated with social assistive robot to navigate in a physical environment intelligently [1]. In the experiments, the *Zenbo Patrol* navigated in a room with white lines on the ground as a campus parking lot (Fig. 1(a)), the robot changed angle view of the camera automatically to capture the images around with format of plate numbers, and then recognized the license plate number through zero-shot learning of GPT-4o (Generative Pre-trained Transformer 4 Omni) model in real-time without the need of preprocessing. Compared to the traditional dual-model pipeline combining the YOLO (You Only Look Once) model (YOLOv11n model) and the OCR (Optical Character Recognition) models (Tesseract model and EasyOCR model) in object detection, the license plate recognition in the work uses a single API (Application Programming Interface) to facilitate the development and maintenance processes [2]. The recognized numbers match with the record of the plate numbers in the system, and notifies the system manager via Line chatbot (Line messaging API) if an illegal parking is detected.

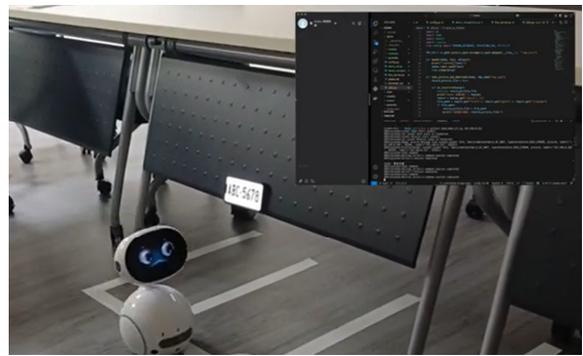

Fig. 1. *Zenbo Patrol* in an indoor simulated parking lot scenario (ASUS Zenbo Junior II).

## 2. RELATED WORKS

Many researches used the dual-model pipeline combining object detection model and the OCR model in the research filed of ALPR (Automatic License Plate Recognition). YOLO model was adopted to detect the region of license plates in an image, and then recognizes texts on the cropped image through the Tesseract model or the EasyOCR (Easy Optical Character Recognition) model [3]. The YOLOv3-tiny model also achieved 96.9% accuracy across eight datasets [4]. Although the dual-model pipeline method performed well in a controlled environment, the complex outdoor environments such as low light, strong reflection or motion blur affected the object detection, and the angle changes of license plates and low resolution data also have influences on the text recognition, and the data preprocessing includes the data collection, training and annotation in both the detection model and the OCR model. In addition, for license plates with different formats or special fonts across countries, current ALPR systems using dual-model pipeline method usually designed for specific location and reduce the capability and flexibility in license plate recognition.

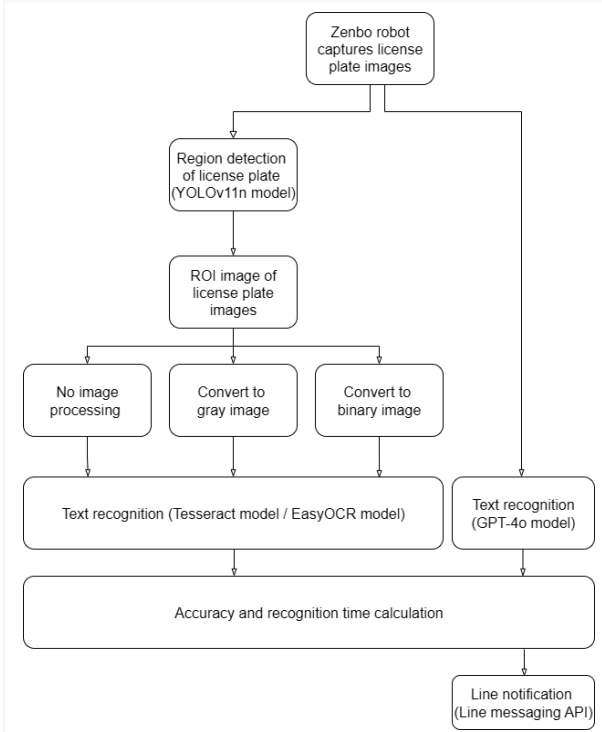

Fig. 2. Comparison of the dual-model pipeline method and LLMs in the system design of the *Zenbo Patrol*.

For the advantages of image understanding and text recognition of LLMs, semantic data outperformed the traditional OCR model in the OCRBench dataset, and still has the limitation in the recognition of random character sequences [5], and the VLMs (Vision Language Models) model interprets images with text explanations. Google company announced a series of PaLI (Pathways Language and Image Model) models (PaLI, PaLI-X, PaLI-3) for training multilingual image-text data such as TextVQA dataset and scene-text understanding, and demonstrated the strong cross-lingual ability [6]. Furthermore, the Kosmos-2.5 model of Google company enhances the data process of both text blocks and markdown structures simultaneously. LLaVA-NeXT model (Large Language and Vision Assistant-Next Generation) improves recognition and interpretation using an expanded visual instruction dataset with high resolution input. In addition, LLaVA-Read model combines dual vision encoders with layout-aware pretraining, and performed well close to the GPT-4V (Generative Pre-trained Transformer 4 with Vision) model in the OCRBench dataset [7].

For the non-semantic and variable properties of plate numbers, LMMs is gradually replaced the traditional dual-model pipeline method for the interpretation of multiple data in license plate recognition. OneShotLP (One-Shot License Plate) model combines the GPT-4V model and the SAM (Segment Anything Model) model to achieve 98.5% F1 score in a real video dataset such as the UFPR-ALPR dataset and the SSIG-SegPlate dataset without additional training data [8]. VehiclePaliGemma proposed a lightweight cloud ALPR model to fine-tuned the Malaysian license plate dataset, and outperformed the GPT-4o model [9]. In addition, LMMs has high performance in license plate recognition by pretraining of large-scale image-text data, GPT-4 model performed well in complex lighting and oblique angle conditions, and the PaLI-Gemma model achieved high accuracy using fined-tuned small dataset [9,10]. Hence, in the study, the GPT-4o model was adopted for license plate recognition in real-time, and integrated with a social assistive robot for using in a real scenario.

## 3. SYSTEM DESIGN

The *Zenbo Patrol* has developed for license plate recognition and illegal parking notification, and the dual-model pipeline contains license plate region (YOLOv11n model) and text recognition (Tesseract model and EasyOCR model), and the LLMs of the GPT-4o model for image understanding were compared (Fig. 2). Real-time recognition and flexibility of the model were considered in the system design of the *Zenbo Patrol*.

### 3.1. Evaluation of license plate recognition

License plate dataset in Taiwan was used in license plate recognition (50 color images; different resolutions: width (2114 to 3264), height (2114 to 2994)) [11]. Average recognition accuracy and the average recognition time are used for quantitative evaluation of each model. The information of the license plate number was annotated manually on the file name of each image for matching the recognition results of each model, and the average accuracy of the total 50 images are calculated (implemented in Python language). For calculation the average recognition time of each model, the timestamp was record from the start point and the end point in the processing of each image, and the average of recognition time of the total 50 images are calculated in seconds. In addition, the steps of image load, interpretation and result return were also calculated in the GPT-4o model.

Table 1. Comparison of the dual-model pipeline method and LLMs in license plate recognition.

| Models | Average recognition accuracy (%) of original image | Average recognition accuracy (%) of original image (ROI image) | Average recognition accuracy (%) of gray image (ROI image) | Average recognition accuracy (%) of binary image (ROI image) | Average recognition time (sec.) |
|---|---|---|---|---|---|
| YOLOv11n + Tesseract | X | 54 | 58 | 60 | 0.6963 |
| YOLOv11n +EasyOCR | X | 70 | 72 | 50 | 0.3937 |
| GPT-4o | 98 | X | X | X | 2.8048 |

*3.1.1. Combination of YOLO model and OCR model in dual-model pipeline method*

The dual-model pipeline method contains license plate region (YOLOv11n model) and text recognition (Tesseract model and EasyOCR model). From an input of an image, a pretrained YOLOv11n model (Training set: 85.69%, validation set: 14.31%, precision (P) = 0.794, Recall(R) = 0.997, mean Average Precision (mAP50) = 0.994, Intersection over Union (IoU) = 0.9236) was adopted for detecting the region of a license plate, and the ROI (Region of Interest) image contains the bounding box of the license plate. The ROI image (Fig. 3(a)) is converted to gray image (Fig. 3(b)) for reducing background noises and improving processing speed, and then is converted to a binary image (Fig. 3(c)) through thresholding algorithm of Otsu for enhancing contrast between text and the background [12]. Tesseract model and EasyOCR model were adopted in text recognition, the former one is an open-source optical character recognition model of Google company with multilingual support, character training, contextual recognition and automatic layout analysis, and the latter one is a deep learning model combining the CRAFT (Character Region Awareness for Text Detection) model and the CRNN (Convolutional Recurrent Neural Networks) model with capabilities of text in natural scenes and irregular fonts.

From the comparison results of the ROI image, gray image and binary image of each model (Table 1), the gray image (Accuracy: 58%) and the binary image (Accuracy: 60%) performed close to the original image (Accuracy: 54%) in the Tesseract model, and reveals low sensitivity of the model in intensity variations. Although the performance of the binary image (Accuracy: 50%) in the EasyOCR model is poor, the

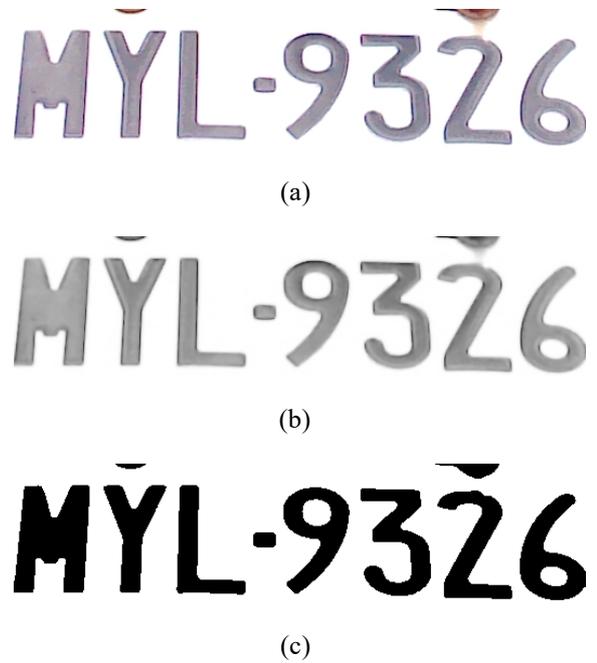

(a)

(b)

(c)

Fig. 3. Prepressing of an ROI image of a license plate image: (a) original image, (b) gray image and (c) binary image.

original image (Accuracy: 70%) and the gray image (Accuracy: 72%) in the EasyOCR model performed better than that of the Tesseract model, and shows capability of intensity variations in the EasyOCR model. However, after image preprocessing, the noises on the images such as stains, reflections or blurred edges still have influences on the performance of the Tesseract model and the EasyOCR model. Sometimes, the low accuracies occurred on the images after preprocessing that the preprocessing may increase image noises. From the results of the average recognition time of each model (Tabel 1), the EasyOCR model performed best

(Time: 0.3937s), and the Tesseract model almost spent twice time (Time: 0.6963s) than the EasyOCR model.

*3.1.2. GPT-4o large multimodal model*

GPT-4o model, a LLMs model was adopted for understanding a license plate image without the preprocessing of the region detection in the dual-model pipeline method, and processes textual and image data simultaneously to generate an interpretation. From an input of an original license plate image, the prompt input of the GPT-4o model is in natural language as follows: "This image is a photo of a car or motorcycle license plate. Please output only the license plate number shown in the image, in a format of 6 or 7 characters composed of English letters and numbers, such as "ABC1234." The license plate number should retain only alphanumeric content, with all spaces and hyphens removed. Do not add any annotations or extra text, only return the license plate number." The output of an interpretation through the zero-shot learning of the GPT-4o model is in a customized format such as "Process: HPJ149.jpg, Correct: HPJ149, Prediction: HPJ149, If Correct: True, Accurate recognition: 6/6, Accuracy: 100%."

From the results of the original image in the GPT-4o model (Table 1), the model performed best (Accuracy: 98%) than the other models, and reveals better noise tolerance of intensity variations and angle changes. From the results of the average recognition time in the GPT-4o model (Tabel 1), the model spent more time (Time: 2.8048s) than the other models. The delay happened on the repeat of the API call (>7), and the API response time is tolerable in the *Zenbo Patrol* application for the high accuracy recognition. Hence, the LLMs performs better than the dual-model pipeline method in license plate recognition, and the advantage of the LLMs is without any preprocessing of the images. In the study, the GPT-4o model was adopted in license plate recognition.

### 3.2. Interface design

The interface design of the *Zenbo Patrol* is in the form of a Line message. The recognition time of the cloud GPT-4o model is tolerable about 2 or 3 seconds. After license plate recognition, the robot matches the license plate number and sends the information of the recognition via Line chatbot (Line messaging API) in real-time. In a campus scenario, the robot sends the information of license plate number, captured time and the place to the system manager when a legal parking is detected (Fig. 4 (a)), and if an illegal parking is detected, the robot sends the same information to the system manager with a warning (Fig. 4 (b)). Hence, the Line messages help the system manager to check the parking status immediately, and the providing information is also record in the system.

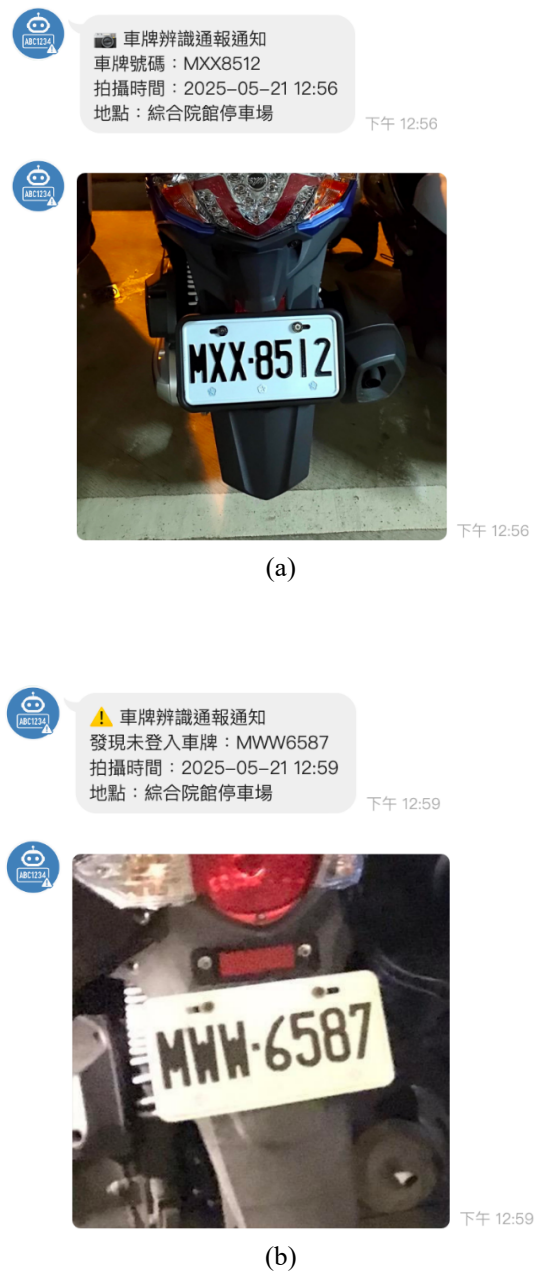

Fig. 4. Line message notification: (a) legal parking notification, (b) illegally parking

### 4. CONCLUSIONS

*Zenbo Patrol* has developed for solving the illegal parking problem in a real scenario, and has the functions of license plate recognition and Line message notification in real-time. Large multimodal model was adopted and performed well with high accuracy in the experiments. In the future work, multiple license plates or image noises of vibrations can be also considered. *Zenbo Patrol* can be applied in a private or a public indoor environment.

## ACKNOWLEDGEMENT

This work was partially supported by the National Science and Technology Council, Taiwan, under grants 114-2635-E-004 -001-.